\documentclass[letterpaper, 10 pt, journal, twoside]{IEEEtran}
\ifCLASSINFOpdf
  \usepackage[pdftex]{graphicx}
\else
\fi
\usepackage{amsmath}
\usepackage{amsfonts}
\usepackage{booktabs}
\usepackage{tabularx}
\usepackage{multirow}
\usepackage[table]{xcolor}
\usepackage{colortbl}
\usepackage{makecell}
\ifCLASSOPTIONcompsoc
 \usepackage[caption=false,font=normalsize,labelfont=sf,textfont=sf]{subfig}
\else
 \usepackage[caption=false,font=footnotesize]{subfig}
\fi
\begin{document}

%
\title{Vector Field Augmented Differentiable Policy Learning for Vision-Based Drone Racing}
%
%
%

\author{Yang Su, Feng Yu, Yu Hu, Xinze Niu, Linzuo Zhang, Fangyu Sun, and Danping Zou%
\thanks{Manuscript received: November 29, 2025; Revised: February 3, 2026; Accepted: March 3, 2026.}
\thanks{This paper was recommended for publication by Editor Aniket Bera upon evaluation of the Associate Editor and Reviewers' comments. This work was supported
by National Key R\&D Program of China (2022YFB3903802) and National
Science Foundation of China (62073214). \textit{(Yang Su and Feng Yu contributed equally to this work.)} \textit{(Corresponding author: Danping Zou.)}} 
\thanks{The authors are with the Shanghai Jiao Tong University, Shanghai 200240,
China (e-mail: dpzou@sjtu.edu.cn).}%
\thanks{Digital Object Identifier (DOI): see top of this page.}
}
%
%

\markboth{IEEE Robotics and Automation Letters. Preprint Version. Accepted MARCH, 2026}
{Su \MakeLowercase{\textit{et al.}}: Vector Field Augmented Differentiable Policy Learning for Vision-Based Drone Racing} 

%



\maketitle

\begin{abstract}

Autonomous drone racing in complex environments requires agile, high-speed flight while maintaining reliable obstacle avoidance. Differentiable-physics-based policy learning has recently demonstrated high sample efficiency and remarkable performance across various tasks, including agile drone flight and quadruped locomotion. However, applying such methods to drone racing remains difficult, as key objectives like gate traversal are inherently hard to express as smooth, differentiable losses. To address these challenges, we propose DiffRacing, a novel vector field–augmented differentiable policy learning framework. DiffRacing integrates differentiable losses and vector fields into the training process to provide continuous and stable gradient signals, balancing obstacle avoidance and high-speed gate traversal. In addition, a differentiable Delta Action Model compensates for dynamics mismatch, enabling efficient sim-to-real transfer without explicit system identification. Extensive simulation and real-world experiments demonstrate that DiffRacing achieves superior sample efficiency, faster convergence, and robust flight performance, thereby demonstrating that vector fields can augment traditional gradient-based policy training with a task-specific geometric prior.

\end{abstract}


%
\IEEEpeerreviewmaketitle

\section{Introduction}

\IEEEPARstart{A}{utonomous} drone flight, particularly in tasks requiring agile and safe navigation through complex environments, remains a central challenge in robotics research. Among these tasks, drone racing exemplifies this challenge by demanding precise control to navigate through gates and avoid obstacles. Classical approaches \cite{han2021fast}, \cite{penicka2022minimum} typically employ a cascaded perception-planning-control pipeline, which, despite achieving impressive performance, suffers from significant computational overhead, error accumulation, and control latency, thereby limiting its practicality in real-world platforms.

In recent years, inspired by the ability of human pilots to learn directly from visual perception, researchers have explored using reinforcement learning for drone racing control from pixels to actions \cite{kaufmann2023champion}, \cite{wang2025environment}, \cite{yu2025mastering}. The method proposed in \cite{yu2025mastering} has addressed the challenge of drone racing in cluttered environments by employing a two-phase reinforcement learning framework—consisting of soft-collision training followed by hard-collision refinement. While effective in achieving robust gate traversal, this approach entails an intricate multi-stage training pipeline and curriculum design, and it inherently suffers from the low sample efficiency characteristic of RL-based methods under sparse rewards. 

\begin{figure}
    \centering
    \includegraphics[width=1\linewidth]{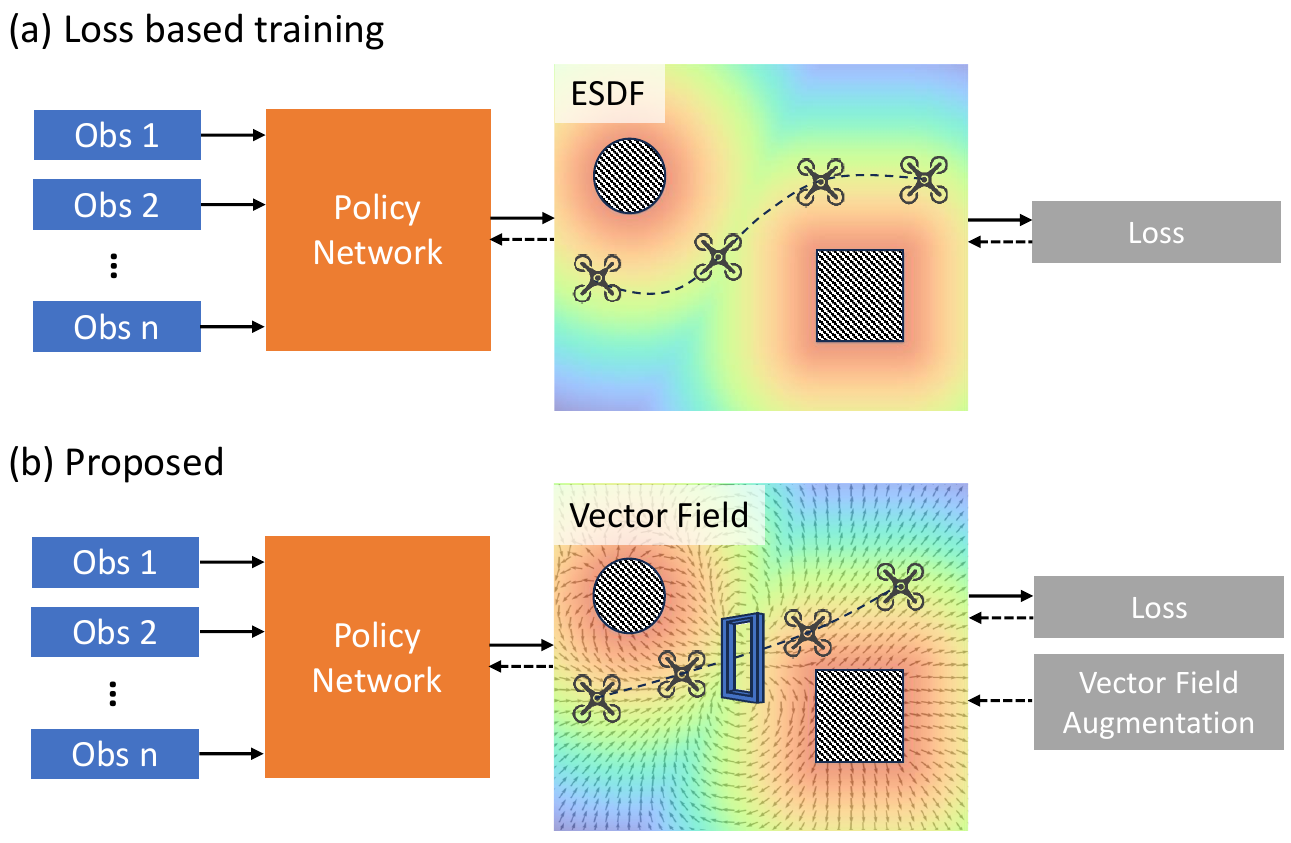}
    \caption{Compared to typical differentiable-dynamics-based methods that rely solely on dense differentiable loss functions to provide gradients, our framework integrates Attractive Vector Fields as a \emph{geometric prior} for gate traversal, alongside standard loss functions to ensure safety.}
    \label{method_comparison}
\end{figure}

Alternatively, differentiable-dynamics-based methods have recently gained attention for their high sample-efficiency and remarkable real-world performance \cite{zhang2025learning, song2024learning,heeg2025learning,schwarke2025learning}. Policies can be trained via backpropagation through time (BPTT) \cite{mozer2013focused}, directly propagating loss gradients through the simplified system dynamics. This approach provides more accurate analytic gradients and significantly improves sample efficiency. However, it relies heavily on dense differentiable loss functions defined over differentiable robot states—for example, penalizing proximity to obstacles to ensure safety, as illustrated in Fig. \ref{method_comparison}(a). Designing a fully differentiable loss for racing tasks is particularly challenging.
For example, gate-passing performance is naturally defined as a binary success indicator—either the drone crosses the gate or it does not. Such an objective is inherently non-differentiable, and straightforward smooth approximations often lead to conflicting objectives between safety and racing goals. As a result, differentiable methods easily become trapped in local optima or exhibit overshooting behavior during optimization, especially at high speeds. Therefore, directly extending differentiable training pipelines to complex racing scenarios remains difficult.

\begin{figure*}[htbp]
    \centering
    \includegraphics[width=0.95\linewidth]{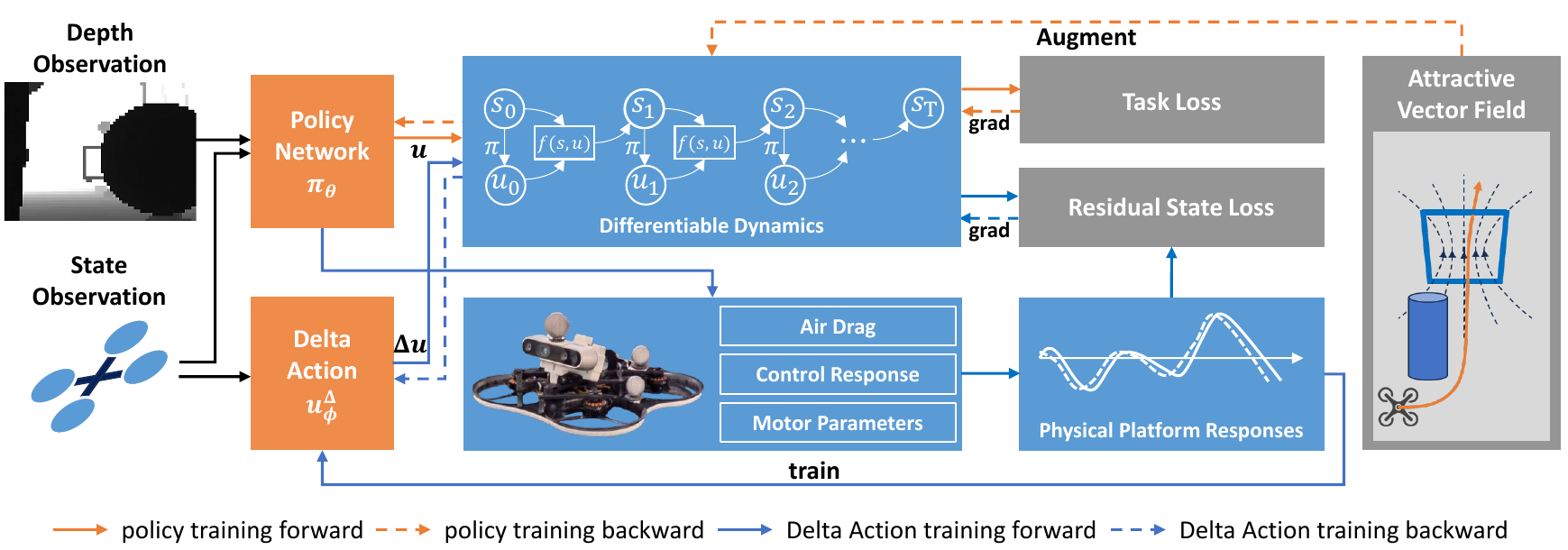}
    \caption{Overview of the DiffRacing Framework. The policy network ($\pi$) takes depth and state observations as input. The core components are: the differentiable simulator, which allows loss gradients to back-propagate to the network; the Attractive Vector Field module, which augments policy gradients during training. The detailed augmentation mechanism is provided in Sec.~\ref{sec:vector_field_aug_policy_learning}; and the Delta Action Model, which is trained to compensate for dynamics mismatch. Dashed lines indicate gradient flow. The orange lines represent the data flow during policy training, where the entire pipeline—including UAV dynamics rollout, image rendering, loss computation, and gradient backpropagation—is carried out fully within the simulator. The blue lines represent the data flow described in Sec.~\ref{sec:delta_action}, which includes data collection, Delta Action Model training, and subsequent policy fine-tuning.}
    \label{framework}
\end{figure*}

Inspired by the physical magnetic field, we leverage the intuition that a closed current-carrying loop can generate magnetic field lines that thread through the loop itself, yielding a strong geometric prior for “through-the-gate” traversal. We therefore construct Attractive Vector Fields (AVF) based on the analytic magnetic-field formulation. We further develop a novel framework which integrates Attractive Vector Fields (AVF) into the differentiable training pipeline, as shown in Fig. \ref{method_comparison}(b). Rather than directly commanding the drone’s velocity to follow the vector field, our approach constructs a composite vector field by combining gradients induced by differentiable loss functions with Attractive Vector Fields during training. Furthermore, to achieve agile and robust racing performance, we incorporate the Delta Action Model proposed in \cite{he2025asap}—which surpasses conventional System Identification and Delta-Dynamics methods—into our differentiable learning framework. In contrast to \cite{he2025asap}, our approach trains the Delta Action Model using analytic gradients from the differentiable simulator rather than PPO \cite{schulman2017proximal}, enabling faster convergence. We validate the effectiveness of our fully differentiable framework through extensive experiments in both simulation and real-world platforms. In a sim-to-sim comparison study with the state-of-the-art approach proposed in \cite{yu2025mastering}, which relies on a multi-phase training strategy and curriculum learning to handle cluttered environments, our method demonstrates that a streamlined, single-stage design can also elegantly solve the trade-off between racing speed and obstacle avoidance while achieving better performance.

Our key contributions are summarized as follows:

\begin{itemize}

     \item We propose a novel differentiable training framework that integrates Attractive Vector Fields as a \emph{geometric prior} during training, enabling the learned policy to achieve more adaptive and dynamic maneuvers.

    \item We incorporate the differentiable Delta Action Model into our sim-to-sim and sim-to-real transfer to compensate for dynamics mismatch and demonstrate its practicability in drone racing tasks.
    \item We validate the proposed framework's effectiveness, robustness, and real-world applicability through extensive experiments in both high-fidelity simulations and real-world deployments.
\end{itemize}

%

\section{Related Work}
In recent years, a series of autonomous drone racing competitions have emerged \cite{moon2019challenges,foehn2022alphapilot,madaan2020airsim}, significantly advancing the capabilities of autonomous drone flight. Approaches have gradually evolved from traditional optimization-based methods \cite{han2021fast, penicka2022minimum, wang2023polynomial} to end-to-end learning \cite{kaufmann2023champion, wang2025environment, doi:10.1126/scirobotics.adg1462}. However, most current methods still focus on high-speed flight in open environments, with relatively few addressing complex and unknown scenarios. Penicka et al. \cite{penicka2022minimum} propose a sample-based method for navigating cluttered environments in minimum time, using incrementally complex quadrotor models to compute time-optimal trajectories. However, it lacks generalization to novel environments. To improve generalizability, Yu et al. \cite{yu2025mastering} introduced an RL-based framework for unknown racing tracks using a two-phase training scheme involving soft-collision and hard-collision refinement, integrated with curriculum learning for better stability. Nevertheless, the complexity of its training architecture and the low sample efficiency caused by sparse gate-crossing rewards remain significant limitations.

The vector field method assigns a velocity or acceleration vector to each point in space, enabling robots to navigate by following the vector at their current position \cite{1087247, 4252175}. Early research primarily aims to extend vector fields to three-dimensional environments and mitigate local optimum problems in obstacle avoidance and path tracking \cite{126315,131810,4282974,5504176}. Building on these efforts, recent studies have introduced rotational or vortex components into artificial potential fields to explicitly break the conservative nature of gradient-based navigation. For instance, recent work \cite{11260944} proposed an enhanced adaptive APF that integrates a 3D vortex force and terminal acceleration, effectively mitigating local minima and oscillations in dynamic environments. Szczepanski \cite{10168227} proposed the Safe Artificial Potential Field (SAPF), which introduces a mechanism to smoothly switch between repulsive and vortex potentials, enabling the robot to bypass local minima, thus effectively mitigating oscillations in narrow corridors. While these advanced potential field methods demonstrate the efficacy of vortex-like structures in avoiding local optima, they are primarily deployed as run-time planners based on analytical force computations. In contrast, our DiffRacing framework leverages vector fields as a geometric prior to augment neural policy learning. 

Differentiable policy learning refers to the use of differentiable simulations to train robot control policies \cite{xu2021accelerated, xing2024stabilizing, song2024learning}. Zhang et al. \cite{zhang2025learning} propose a CUDA-based point-mass model that is directly integrated into policy optimization, enabling neural network control policies to be trained with more accurate analytical gradients and, consequently, improving sample efficiency and convergence speed. Improving on \cite{zhang2025learning}, Lee et al. \cite{lee2025quadrotor} incorporate a privileged global time-of-arrival (TOA) map, enabling supervision that facilitates more globally effective policy optimization. While the TOA map can be interpreted as a form of field, the resulting slow flight speeds further limit its effectiveness in drone racing scenarios.

\section{Methodology}
\subsection{Overview} 

The proposed framework leverages the high sample efficiency of differentiable training and employs vector fields to augment the standard gradient-based policy learning, thereby resolving conflicts between loop-passing and obstacle-avoidance optimization. As illustrated in Fig. \ref{framework}, our framework comprises four main components: (1) a differentiable dynamics simulator (Sec.~\ref{sec:diff_dynamics}) that allows gradients of differentiable states to back-propagate to the policy and Delta Action Model network; (2) a vector field augmentation module (Sec.~\ref{sec:vector_field_aug_policy_learning}) that utilizes Attractive Vector Fields to augment policy learning during training; (3) a Delta Action Model (Sec.~\ref{sec:delta_action}) that outputs a correction in the action space to compensate for dynamics mismatch and facilitates sim-to-real transfer; (4) a policy network (Sec.~\ref{sec:training_details}) that takes depth observations and state observations as input and outputs acceleration commands.

\subsection{Differentiable Dynamics} \label{sec:diff_dynamics}

Differentiable dynamics-based methods formulate the drone control task as a Markov decision process, where system dynamics are modeled as \(\boldsymbol{s}_{k+1}=f(\boldsymbol{s}_k,\boldsymbol{u}_k)\). Here \(\boldsymbol{s}_k\) denotes the system state at step $k$, including position, orientation, velocity, etc. The control command \(\boldsymbol{u}_k\), representing the desired body frame acceleration, is generated by a deterministic policy \(\boldsymbol{u}_k=\pi_\theta(\boldsymbol{o}_k)\), parameterized by \(\theta\) and taking the system observation \(\boldsymbol{o}_k\) as input. 

\begin{equation}
    \boldsymbol{o}_k=[^b\boldsymbol v_k, \mathbf{r_3}, ^b\boldsymbol{p}_k^{\text{gate}},\boldsymbol{u}_{k-1}, \boldsymbol{I}_{k}]
    \label{observations}
\end{equation}

Here, \(^b\boldsymbol v_k\) represents the velocity in the body frame, and \(\mathbf{r_3}\) denotes the transposed third column of the rotation matrix \(\boldsymbol R\). \(^b\boldsymbol{p}_k^{\text{gate}}\) denotes the relative position of the gate in the body frame, and \(\boldsymbol{u}_{k-1}\) is the previous control command. \(\boldsymbol{I}_{k}\) represents the depth image with a resolution of \(24\times32\) \cite{zhang2025learning}.

At each step, loss term \(l_k\) is calculated using the corresponding differentiable loss function \(l_k = l(\boldsymbol{s}_k,\boldsymbol{u}_k)\). The overall loss \(L\) over a finite horizon \(T\) is then defined as the average of these per-step losses: \(L=\frac{1}{T}\sum_{k=1}^{T}l_k\). Given that both the dynamic function \(f(\boldsymbol{s}_k,\boldsymbol{u}_k)\) and the loss function \(l(\boldsymbol{s}_k,\boldsymbol{u}_k)\) are differentiable, the gradient of \(L\) with respect to the policy parameters \(\theta\) can be expressed as:

\begin{equation}
    \frac{1}{T}\sum_{k=0}^{T-1}\left( 
    \sum_{i=0}^{k}
    \frac{\partial l_k}{\partial \mathbf{s}_k}
    \prod_{j=i+1}^{k}\left(
    \frac{\partial \mathbf{s}_j}{\partial \mathbf{s}_{j-1}}
     e^{-\alpha\Delta t}\right)
    \frac{\partial \mathbf{s}_i}{\partial \theta}
    + 
    \frac{\partial l_k}{\partial \mathbf{u}_k}
    \frac{\partial \mathbf{u}_k}{\partial \theta}
    \right)
\end{equation}

where \(\alpha\) is a gradient decay factor proposed in \cite{zhang2025learning} to mitigate gradient accumulation.

\subsection{Vector Field Augmented Differentiable Policy Learning} \label{sec:vector_field_aug_policy_learning}
\subsubsection{Motivation}

Consider the differentiable avoidance loss \(L_{\text{C}} = L_{\text{clearance}} + L_{\text{collide}}\) proposed by \cite{zhang2025learning} as an example, where \(L_{\text{clearance}}\) and \(L_{\text{collide}}\) are functions of the position state \(\mathbf{p}\):
\begin{equation}
    L_{\text{clearance}} = \frac{1}{T}\sum_{k=1}^T \beta_1 \ln(1 + e^{\beta_2 (||\mathbf{d}_k|| - r_q) })
\end{equation}

\begin{equation}
    L_{\text{collide}} = \frac{1}{T}\sum_{k=1}^T v_k^c \max(1 - (||\mathbf{d}_k|| - r_q ), 0)^2
\end{equation}
Here, \(\mathbf{d}_k = \mathbf{p}_k^{\text{nearest}} - \mathbf{p}_k\) denotes the relative position of the closest obstacle, \(r_q\) is the safety radius of the drone, and \(v_k^c\) is the speed toward the closest obstacle. \(\beta_1\) and \(\beta_2\) are weighting coefficients. Notably, \(v_k^c\) is used to scale the loss function and is detached from the computation graph, making it gradient-free. By minimizing the loss function \(L_{\text{C}}\), the negative gradient \(-\nabla_{\mathbf{p}} L_C\) with respect to the position \(\mathbf{p}\) is automatically calculated. This gradient can be interpreted as an induced vector field acting on the position state, which is back-propagated to the network parameters \(\theta\). Consequently, the policy is guided away from obstacles by following the direction of the induced vector field during training.

While these differentiable loss functions are effective for simple obstacle avoidance, they tend to create local optima in drone racing scenarios. For example, when the gradient from the collision loss (repelling the drone from the gate frame) and the gate-passing loss (attracting it to the gate center) act in opposite directions, these conflicting gradients may cancel each other out or create a saddle point, resulting in a vanishing gradient that hinders effective navigation. Inspired by \cite{panagou2014motion}, the vortex-based vector field they proposed introduces rotational components into the guidance field, helping to mitigate the issue of local optima. Therefore, in addition to the irrotational (curl-free, \(\nabla \times \nabla_{\mathbf{p}} L_C = 0\)) vector fields induced by differentiable loss functions, we incorporate a novel rotational Attractive Vector Field \(\mathbf{u}_A\) (\(\nabla \times \mathbf{u_A} \neq 0\)) into the differentiable training framework. This field is combined with the previously discussed gradient field $-\nabla_{\mathbf{p}} L_C$ to form a composite guidance signal on the position state:
\begin{equation}
    \mathbf{u}=\mathbf{u}_A-\nabla_\mathbf{p} L_C
    \label{avf_naive_integration}
\end{equation}

\begin{figure}[htbp]
    \centering
    \subfloat[]{
        \includegraphics[width=0.48\linewidth]{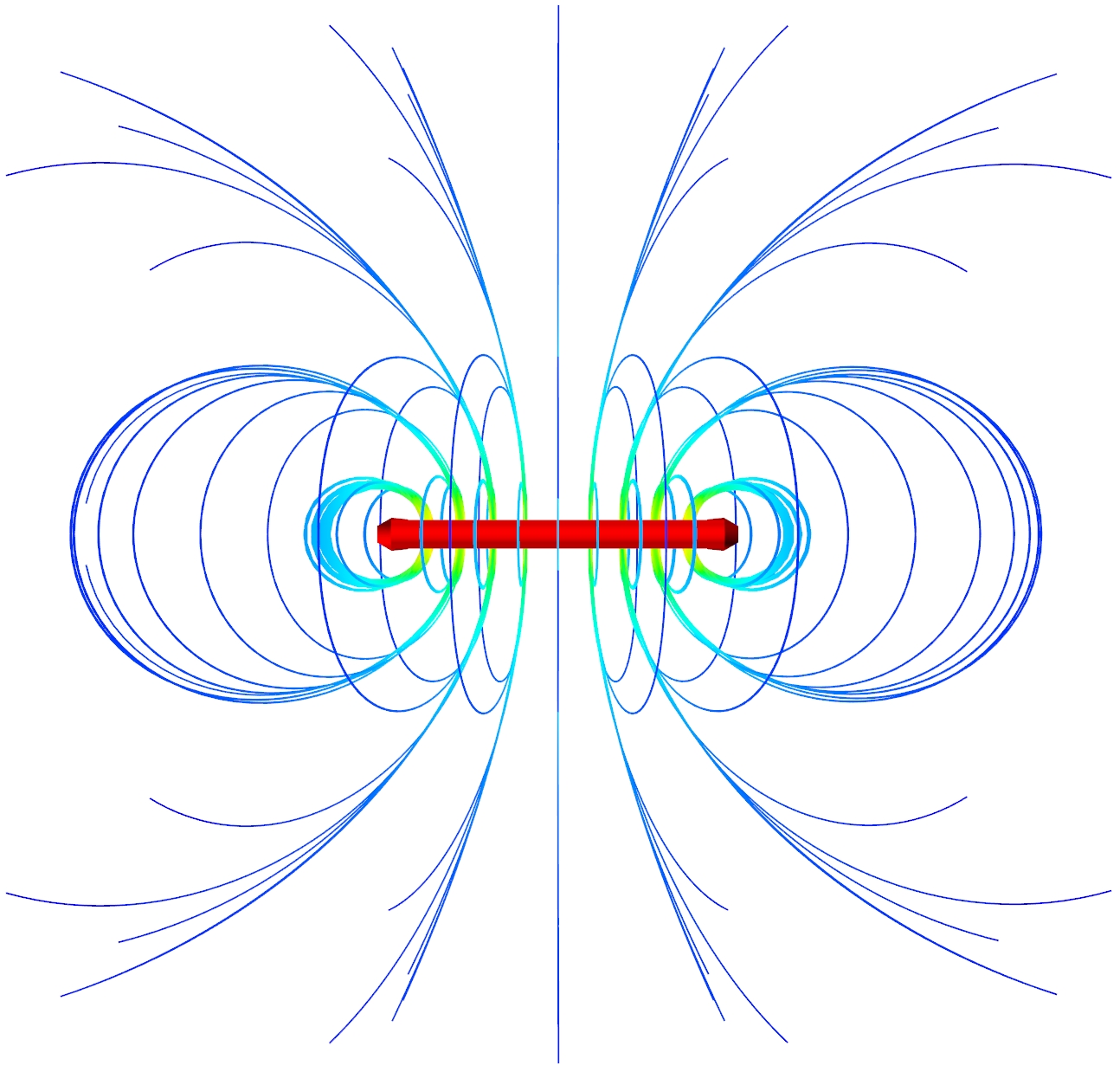}
        \label{fig:mag_2d_horizontal}
    }%
    \hfil
    \subfloat[]{
        \includegraphics[width=0.46\linewidth]{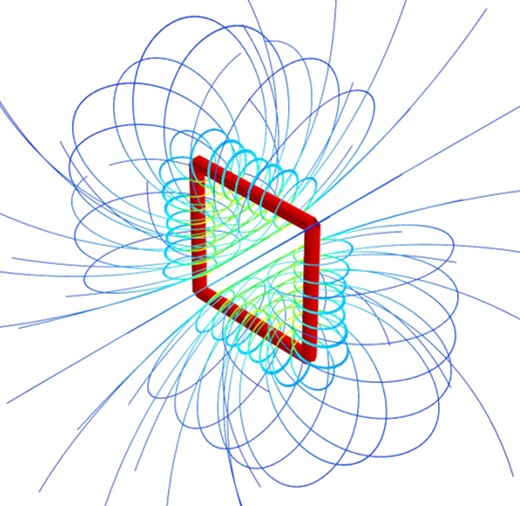}
        \label{fig:mag_3d_horizontal}
    }
    \caption{3D Magnetic field visualization: (a) Top view of magnetic field; (b) Axonometric view of magnetic field.}
    \label{mag_fields_horizontal}
\end{figure}

\subsubsection{Attractive Vector Fields} \label{sec:vector_field_construction}

\begin{figure*}[t!]
    \centering

    \includegraphics[width=0.95\linewidth]{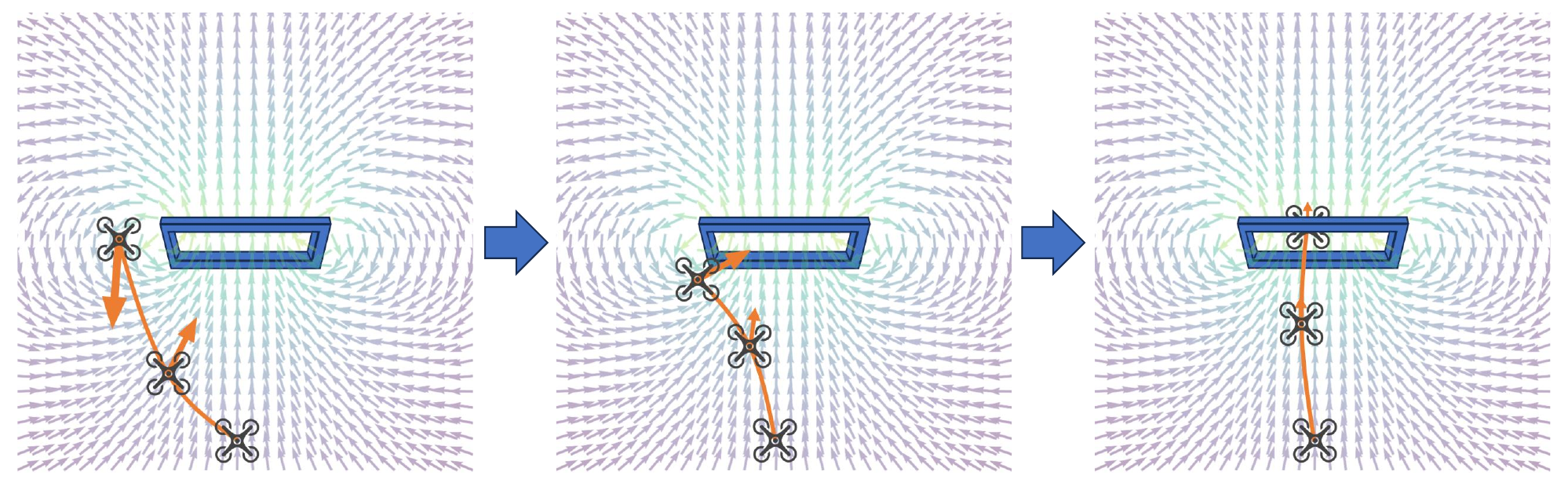}
    \caption{An Intuitive Illustration of \emph{Trajectory Self-Correction} (Top View): From left to right, an overshooting trajectory is gradually refined under the continuous guidance of the Attractive Vector Fields.}
    \label{self_correction}
\end{figure*}

{Our Attractive Vector Field formulation leverages the intuition from physics: the magnetic field of a closed current-carrying loop forms field lines that thread the loop itself, providing a strong geometric prior for “through-the-gate” traversal. We first construct a three-dimensional magnetic vector field induced by the gate}, where each gate is modeled as a rectangular current loop composed of four wire segments, as illustrated in Fig. \ref{mag_fields_horizontal}.
For a spatial point $\mathbf{p}$, the magnetic field generated by the \(i\)-th segment is given by the simplified Biot–Savart law:
\begin{equation}
\mathbf{B}_i(\mathbf{p})
= C_I\cdot\hat{\mathbf{l}}_i
\left(
\frac{\mathbf{r}_{i1}-\mathbf{p}}{\|\mathbf{r}_{i1}-\mathbf{p}\|}
- \frac{\mathbf{r}_{i2}-\mathbf{p}}{\|\mathbf{r}_{i2}-\mathbf{p}\|}
\right)
\frac{\hat{\mathbf{l}}_i \times \mathbf{d}_i}{\|\mathbf{d}_i\|^2}
\label{eq:mag_segment}
\end{equation}
Here, $\mathbf{r}_{i1}$ and $\mathbf{r}_{i2}$ are the positions of the start and end points of the $i$-th wire segment, 
$\hat{\mathbf{l}}_i = {(\mathbf{r}_{i2}-\mathbf{r}_{i1})}/{\Vert\mathbf{r}_{i2}-\mathbf{r}_{i1}\Vert}$ 
is its unit direction vector, 
and $\mathbf{d}_i$ 
is the shortest vector from the query point $\mathbf{p}$ to the wire segment. 
The constant $C_I$ controls the field strength and is set to \(2\times10^{-5}\) during training, and 
$\hat{\mathbf{l}}_i \times \mathbf{d}_i$ 
determines the direction of the field. The total gate-induced magnetic field is the superposition of the four wire contributions:
\begin{equation}
\mathbf{B}(\mathbf{p}) = \sum_{i=1}^{4} \mathbf{B}_i(\mathbf{p})
\label{eq:mag_total}
\end{equation}
This formulation yields a divergence-free (solenoidal) magnetic field that forms an inward swirling flow, 
naturally guiding the drone toward and through the gate center. 

Finally, the Attractive Vector Field \(\mathbf{u}_A\) is defined as:
\begin{equation}
\mathbf{u}_A = \left(1 - \frac{\mathbf{v}\cdot \mathbf{B}(\mathbf{p})}{\Vert \mathbf{v} \Vert \Vert \mathbf{B}(\mathbf{p}) \Vert+\epsilon}\right)
\frac{\mathbf{B}(\mathbf{p})}{\Vert \mathbf{B}(\mathbf{p}) \Vert^{\lambda_A}}, 0 < \lambda_A < 1
\label{u_a_cal}
\end{equation}
The parameter \(\lambda_A\) controls the rate of decay with distance and is set to 0.3 during training, and \(\epsilon\) is set to \(10^{-5}\). This formulation ensures that  \(\mathbf{u}_A\) decays when the agent is already aligned and moving in the desired direction or is far away from the gates. Consequently, it prevents the Attractive Vector Fields from exerting excessive influence on the agent, thereby enhancing training stability and overall safety. Although in this work \(\mathbf{u}_A\) is formulated based on the rectangular gate geometry, the underlying magnetic field calculation principle is easily generalizable to common non-rectangular gate shapes. For example, a circular gate can be approximated by discretizing its circumference into a polygon of straight segments. The total attractive field is then calculated by the superposition of the fields generated by these segments.

\subsubsection{Integration into Differentiable Training}

To effectively balance the conflicting objectives of obstacle avoidance and high-speed gate traversal, we augment the standard gradient-based policy learning with the proposed Attractive Vector Field. The overall loss is defined as a weighted sum of four terms: 
\begin{equation}
    L=\lambda_CL_C+\lambda_aL_a+\lambda_jL_j+\lambda_pL_p
\end{equation}
where $L_C$, ($L_a, L_j$), and $L_p$ correspond to obstacle avoidance, control smoothness, and racing speed performance, respectively. The associated weighting coefficients are set to \(\lambda_C = 3.0, \lambda_a=10^{-2}, \lambda_j = 10^{-3}, \lambda_p = 0.4\).

Specifically, the control smoothness loss terms \(L_a\) and \(L_j\) are defined as: 

\begin{equation}L_a=\frac{1}{T}\sum_{k=1}^T\Vert \boldsymbol{a}_k \Vert^2, L_j=\frac{1}{T-1}\sum_{k=1}^{T-1}\left \Vert \frac{\boldsymbol{a}_{k+1}-\boldsymbol{a}_k}{\Delta t}\right \Vert^2\ \end{equation} 
Here, $\mathbf{a}_k$ denotes the acceleration. The racing speed performance loss \(L_p\), which promotes high-speed gate traversal, is defined as:
\begin{equation}L_p=-\frac{1}{T}\sum_{k=1}^T\frac{^b\boldsymbol{v}_k \cdot ^b\boldsymbol{p}_k^{\text{gate}}}{\Vert ^b\boldsymbol{p}_k^{\text{gate}}\Vert} \label{loss_progress} \end{equation}

This term \(L_p\) explicitly promotes velocity aligned with the target gate direction, ensuring sustained forward progress during racing. Moreover, since the Attractive Vector Field \(u_A\) naturally attenuates at long distances, \(L_p\) provides the necessary long-range directional guidance. Together, they form a complementary guidance mechanism.

Let \(\mathbf{s} = [\mathbf{p}, \mathbf{s}_p]\) denote the complete set of differentiable state variables, consisting of differentiable positions $\mathbf{p}$ and all other differentiable state variables $\mathbf{s}_p$. Then, \(\nabla_{\mathbf{p}}L\) denotes the gradient of the loss function \(L\) with respect to the position variable \(\mathbf{p}\) and \(\nabla_{s_\mathbf{p}}L\) denotes the gradient of the loss function \(L\) with respect to \(\mathbf{s}_p\). During training, \(\nabla_{\mathbf{p}}L\) and \(\nabla_{s_\mathbf{p}}L\) will be back-propagated to the policy parameters \(\theta\) via differentiable dynamics, resulting in the following parameter update rule:

\begin{equation}
\theta \leftarrow \theta - \alpha\left (\nabla_{\mathbf{p}}L \frac{\partial \mathbf{p}}{\partial \theta} + \nabla_{\mathbf{s}_p}L\frac{\partial \mathbf{s}_p}{\partial \theta}\right )
\end{equation}

To incorporate Attractive Vector Fields into the training pipeline, we reformulate the above rule as in Eq. \eqref{avf_naive_integration}, with $\mathbf{u}_A$ defined in Eq. \eqref{u_a_cal}:

\begin{equation}
\theta \leftarrow \theta - \alpha \left [(\nabla_{\mathbf{p}}L-\mathbf{u}_A) \frac{\partial \mathbf{p}}{\partial \theta} + \nabla_{\mathbf{s}_p}L\frac{\partial \mathbf{s}_p}{\partial \theta}\right ]
\label{final_optimization}
\end{equation}
Under this formulation, we augment the standard gradient-based policy learning with \(\mathbf{u}_A\) to provide an explicit geometric prior. Note that the update in Eq. \eqref{final_optimization} is not guaranteed to be the steepest descent direction of the original scalar loss \(L\); instead, we view it as a heuristic yet empirically effective design, as experiments in Sec.~\ref{sec:experiments} demonstrate stable training and outstanding performance.

\subsection{Delta Action Model Learning}\label{sec:delta_action}

In Section \ref{sec:diff_dynamics}, the system dynamics are modeled as a differentiable function \(\mathbf{s}_{k+1} = f(\mathbf{s}_k,\mathbf{u}_k)\). However, achieving a high-fidelity representation of $f$ is challenging in sim-to-real transfer due to real-world factors, such as aerodynamic disturbances and motor response delays. Precisely identifying the dynamics parameters often requires substantial manual effort. To mitigate these model-reality discrepancies and simplify system identification, and motivated by recent work \cite{pan2025learning, he2025asap}, we incorporate Delta Action Model learning into the differentiable learning framework. Following \cite{he2025asap}, we treat the residual dynamics as a Delta Action policy \( u^\Delta_\phi\), which outputs corrections in the action space:

\begin{equation}
    \mathbf{s}_{k+1} = f(\mathbf{s}_k,\mathbf{u}_k+ u^\Delta_\phi(\mathbf{s}_k, \mathbf{u}_k))
\end{equation}

The Delta Action policy \(u^\Delta_\phi\), typically implemented as a neural network, is optimized by minimizing the discrepancy between the recorded real-world trajectory and the corresponding simulated state trajectory under dynamics model $f$:
\begin{equation}
    \min_{\phi}L^\Delta(\phi) = \min_{\phi}\left( \frac{1}{T}\sum_{k=1}^T\Vert \mathbf{s}_k^{\text{sim}}(\phi)- \mathbf{s}_k^{\text{real}}\Vert\right)
\label{delta_action_loss}\end{equation}
where \(\mathbf{s}_k^{\text{sim}}(\phi)\) is the simulated state, obtained by executing the total action \(\mathbf{u}^{\text{real}}_{k-1}+ u^\Delta_\phi(\mathbf{s}^{\text{sim}}_{k-1}, \mathbf{u}^{\text{real}}_{k-1})\) in the simulator, $\mathbf{s}_k^{\text{real}}$ and $\mathbf{u}^{\text{real}}_k$ are the recorded real-world state-action pair. The gradient of loss \(L^\Delta(\phi)\) is back-propagated through the differentiable dynamics model to efficiently update the parameters $\phi$ of the Delta Action policy. This approach effectively leverages the analytical benefits of differentiable simulation to capture dynamics discrepancy, significantly improving sample efficiency and accelerating convergence during training.

In practice, we employ a three-stage pipeline to train and leverage the Delta Action Model. First, an initial dataset of state–action pairs is collected in the target environment using a policy trained without the Delta Action Model. This collected dataset is then used to train the Delta Action Model by minimizing the loss function \(L^\Delta(\phi)\). Finally, the Delta Action Model's parameters are frozen, and its predictions are added to the policy network outputs to form a mixed control action, after which the policy is fine-tuned using the adjusted dynamics.

\subsection{Training Details}\label{sec:training_details}

The policy adopts a compact CNN–RNN architecture to process visual and state information jointly. A CNN encoder extracts spatial features from a \(24 \times 32\) depth image, while a small fully connected branch encodes the 12-dimensional state inputs from the first four terms in Eq. \eqref{observations}. The embeddings are fused and passed through a GRU to capture temporal dependencies. The final cascaded linear layer outputs the acceleration command in the body frame. LeakyReLU ($\alpha=0.05$) is used throughout, and the embedding dimension is set to 192. In contrast to the policy network, the Delta Action policy network excludes the CNN image encoder and is composed only of MLP and GRU modules, with both feature and hidden dimensions reduced to 32. It takes body velocity, orientation and current action as input, and outputs a 3-dimensional action compensation vector.

\section{Experiments}
\label{sec:experiments}

The experimental section validates the core contributions of the proposed DiffRacing framework across four aspects: (1) Ablation studies on the AVF confirm its crucial role in achieving high success rates and speeds. (2) Comparative analysis against PPO and a scalar loss baseline, utilizing the same evaluation metrics, confirms the superior sample efficiency and enhanced robustness of the proposed training scheme. (3) Sim-to-sim evaluations demonstrate the effectiveness of the Delta Action model and its performance advantages over the state-of-the-art baseline. (4) Finally, real-world experiments on a physical drone validate agile and robust flight at speeds up to 6 m/s in unseen, obstacle-dense environments.


\subsection{Ablation on Attractive Vector Fields}
To quantitatively assess the contribution of the AVF, we conducted an ablation study with three training configurations. Performance was evaluated using two metrics: \textbf{Success Cross}, the percentage of trials in which the drone successfully traversed all four gates without collision, and \textbf{Success Rate}, the percentage of trials completed without any collision, regardless of gate traversal. The three experimental configurations are as follows:
(1) W AVF (Ours): The proposed method, performing optimization according to Eq.~\eqref{final_optimization}. (2) W/O AVF, W $L_p$: A baseline trained without the AVF, relying solely on scalar loss functions, especially using the standard progress loss $L_p$ defined in Eq.~\eqref{loss_progress}. (3) W/O AVF, W $L_p^{\mathrm{norm}}$: A baseline trained without the AVF, relying solely on scalar loss functions, especially using the normalized progress loss $L_p^{\mathrm{norm}}$ defined as
\begin{equation}
    L_p^{\mathrm{norm}}=-\frac{1}{T}\sum_{k=1}^T\frac{\left( ^b\boldsymbol{v}_k,^b\boldsymbol{p}_k^{\text{gate}}\right)}{\Vert^b\boldsymbol{v}_k\Vert\Vert ^b\boldsymbol{p}_k^{\text{gate}}\Vert}
\end{equation}
which encourages correct heading toward the gate while ignoring speed optimization.


As shown in Table \ref{tab:performance_comparison_optimized}, the configuration with AVF significantly outperforms both baselines, achieving a 95\% Success Cross and a 97\% Success Rate. In contrast, the baselines struggle to balance speed and safety, either failing to traverse gates effectively (0.0 Success Cross for most configurations) or exhibiting low Success Rate when the progress loss is more heavily weighted to encourage speed. These results highlight the critical role of AVF in providing a continuous geometric prior, enabling the policy to achieve both agile gate traversal and robust obstacle avoidance.

\begin{table}[h!]
\centering
\scriptsize 
\caption{Performance Comparison with and without AVF}
\label{tab:performance_comparison_optimized}
\begin{tabular}{@{}cccccc@{}}
\toprule
\textbf{Settings} & \textbf{$\lambda_p$} & \begin{tabular}[c]{@{}c@{}}\textbf{Success}\\\textbf{Cross}\end{tabular} & \begin{tabular}[c]{@{}c@{}}\textbf{Success}\\\textbf{Rate}\end{tabular} & \begin{tabular}[c]{@{}c@{}}\textbf{Max}\\\textbf{Velocity}\end{tabular} \\
\midrule
W AVF & 0.4 & 0.95 & 0.97 & 6.5 \\
\midrule
\multirow{5}{*}{W/O AVF, W \(L_p\)} & 0.4 & 0.00 & 0.98 & 4.7 \\
& 0.8 & 0.00 & 0.89 & 5.9 \\
& 1.6 & 0.00 & 0.74 & 7.4 \\
& 3.2 & 0.28 & 0.30 & 9.7 \\
\midrule
\multirow{3}{*}{W/O AVF, W \(L_p^{\mathrm{norm}}\)} & 1.6 & 0.00 & 0.95 & 5.0 \\
& 3.2 & 0.00 & 0.84 & 6.1 \\
& 6.4 & 0.58 & 0.58 & 6.3 \\
\bottomrule
\end{tabular}
\end{table}



\subsection{Comparative analysis}

To quantitatively evaluate the performance of different training schemes on the racing task, we compared three settings: (1) PPO \cite{schulman2017proximal}: a standard reinforcement learning baseline; (2) Ours: the proposed framework, leveraging a hybrid gradient derived from a loss function \(L\) and Attractive Vector Field \(\mathbf{u}_A\); and (3) Ours (w/o AVF): A variation of our differentiable policy learning framework where the vector field augmentation is removed. To ensure a fair comparison and prevent the policy from lacking directional guidance entirely, we explicitly added a projection loss term \(L_{\text{proj}}\) into the standard scalar loss function (\(L+\lambda_{\text{proj}}L_{\text{proj}}\)), this term serves as a scalar substitute for the AVF, encouraging the drone to align its heading with the gate:

\begin{equation}
    L_{\text{proj}} = \Vert p_k^{yz|\text{gate}}\Vert^2 \cdot e^{-\beta_3 d_k^{\text{gate}}}
\end{equation}

Here, \(p_k^{yz|\text{gate}}\) represents the position deviation projected onto the gate's \(y-z\) plane, and \(d_k^{\text{gate}}\) denotes the distance to the gate. \(d_k^{\text{gate}}\) is detached from the computation graph and is gradient-free; it is utilized to decay the loss function when the quadrotor is sufficiently distant from the gate. The hyperparameters are set to \(\beta_3=0.5, \lambda_{\text{proj}}=3.0\). This loss term explicitly penalizes deviations from the gate's normal vector, providing a scalar-based directional cue distinct from our vector field approach.

We designed a uniform scalar reward function applied at each time step:
\begin{equation}
    R_k = r_k^{\text{collision}} + r_k^{\text{avoid}}+ r_k^{\text{smooth}} + r_k^{\text{pass}} +r_k^{\text{progress}}
\end{equation}


The detailed expressions of these components are listed in Table \ref{tab:reward_components_reorganized}. Specifically, \(r_k^{\text{pass}}\) is assigned a large positive constant when the quadrotor passes a gate, and remains zero otherwise. Similarly, the collision reward \(r^{\text{collision}}_k\) is set to a negative constant when a collision occurs. The PPO-based method was trained directly using this reward function, while the setting (2) and (3) are evaluated with the same reward function. System dynamics were kept consistent across both methods.

\begin{table}[t]
    \centering
    \scriptsize 
    \caption{Components of the Reward Function}
    \label{tab:reward_components_reorganized}
    \renewcommand{\arraystretch}{1.2}
    \begin{tabular}{ccc}
        \toprule
        \textbf{Term} & \textbf{Expression} & \textbf{Weight ($\lambda$)} \\
        \midrule
        $r_k^{\text{collision}}$ & $\lambda_1 \mathbb{I}(||\mathbf{d}_k||<r_q)$ & $\lambda_1 = -30$ \\[6pt]
        $r_k^{\text{avoid}}$ & \makecell[c]{$\lambda_{\text{2}}\ln{(1+e^{\beta_2(||\mathbf{d}_k||-r_q)})}$ \\  $+\lambda_3v_k^c\max(1-||\mathbf{d}_k||+r_q, 0)^2$} & $\lambda_2 = -6, \lambda_3 = -3$ \\[8pt] 
        $r_k^{\text{smooth}}$ & $\lambda_4 \Vert \mathbf{a}_k\Vert^2 + \lambda_5\left\Vert \frac{\mathbf{a}_{k}-\mathbf{a}_{k-1}}{\Delta t} \right\Vert^2$ & $\lambda_4 = -10^{-2}, \lambda_5 = -10^{-3}$ \\[8pt] 
        $r_k^{\text{pass}}$ & $\lambda_6\mathbb{I}(\Vert ^b\mathbf{p}_k^{\text{gate}}\Vert<r_{\text{th}})$ & $\lambda_6 = 110$ \\[4pt] 
        $r_k^{\text{progress}}$ & $\lambda_7 \frac{\left( ^b\boldsymbol{v}_k,^b\boldsymbol{p}_k^{\text{gate}}\right)}{\Vert ^b\boldsymbol{p}_k^{\text{gate}}\Vert}$ & $\lambda_7 = 0.4$ \\
        \bottomrule
    \end{tabular}
\end{table}

\begin{figure}[t!]
    \centering
    \includegraphics[width=\linewidth]{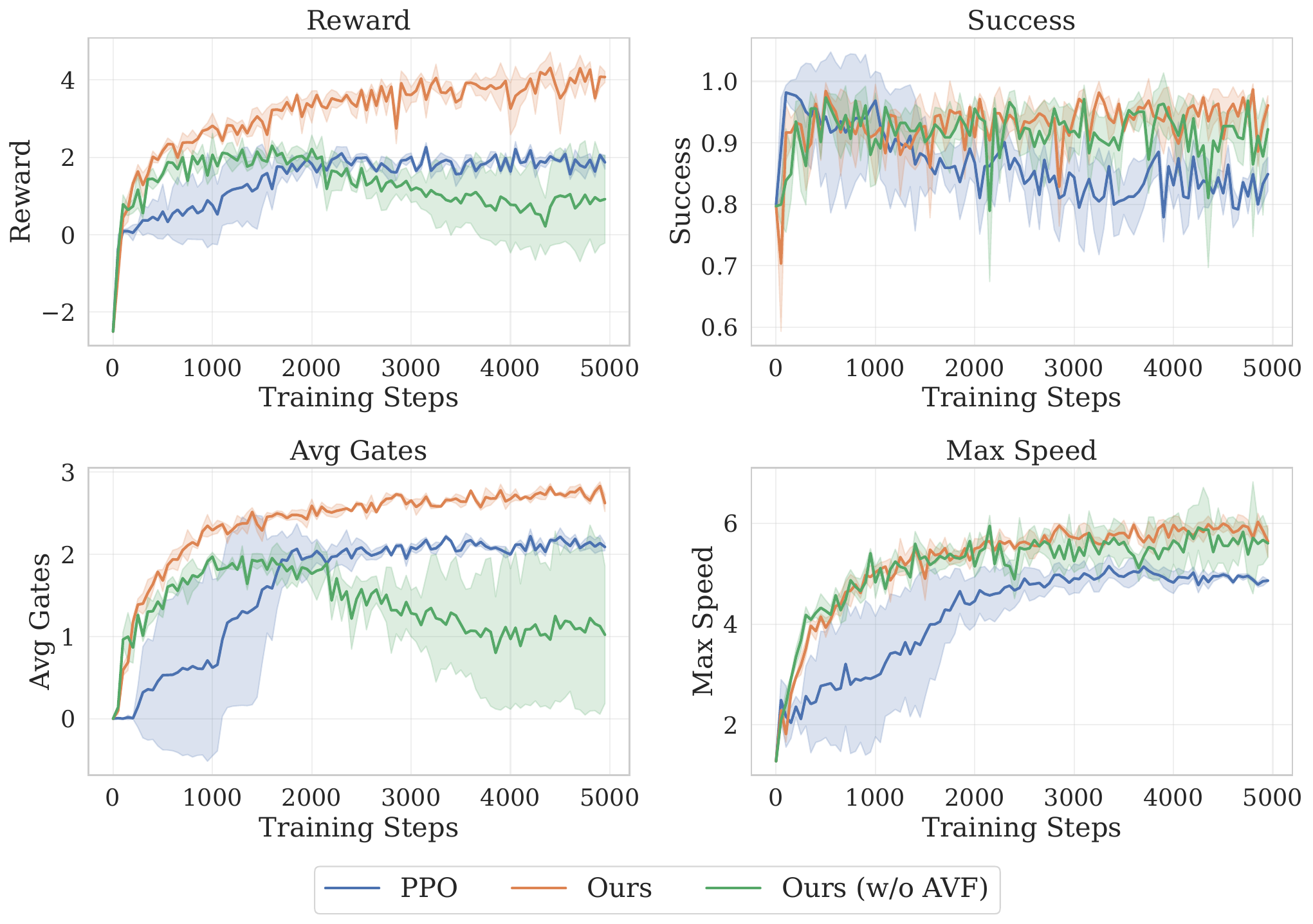}
    \caption{Performance comparison between different training schemes.}
    \label{comparison}
\end{figure}

 As shown in Fig. \ref{comparison}, our method (Ours) with AVF outperforms the PPO-based approach across all key metrics, including Reward, Maximum Speed, Success Rate, and Average Gates Traversed per episode. Notably, our proposed method learned to cross gates effectively from the beginning of training, whereas the PPO-based method initially struggled, exhibiting sudden and unstable learning breakthroughs. This behavior was consistently observed in experiments using a sparse cross reward, where the unpredictable timing of such breakthroughs adversely affected overall training stability. The differentiable baseline (Ours w/o AVF) achieves a high Success Rate, but fails to effectively learn gate crossing ability. This indicates that without the directional guidance of the AVF, the optimization gets trapped in a local optimum where the policy prioritizes safety (avoiding collisions) over the racing objective. These results provide strong evidence that our approach achieves better sample efficiency and more robust optimization performance.

\subsection{Sim-to-Sim Experiments}

We conducted sim-to-sim experiments to evaluate the effectiveness of the Delta Action Model and its performance advantages over the state-of-the-art baseline \cite{yu2025mastering}. The target evaluation platform was IsaacLab \cite{mittal2025isaac}. The policy was trained in a CUDA-accelerated simulator based on \cite{zhang2025learning}, which incorporates differentiable dynamics, and subsequently transferred to IsaacLab for testing. In contrast, the baseline policy in \cite{yu2025mastering} was both trained and tested directly in IsaacLab. Differences in physics engines and controller implementations—Testing in IsaacLab employs a total thrust and body rate (CBTR) controller, whereas training uses a point-mass control model—increase the risk of failure during transfer. To mitigate this, we integrated Delta Action Model learning into the sim-to-sim transfer process and assessed its impact on success rate and velocity performance. Policies were tested across various track geometries (Zigzag, Circular, and Ellipse) and terrain difficulties.

The experimental results, summarized in Table \ref{sim_performance}, demonstrate that DiffRacing with the Delta Action Model (DA) achieves superior flight speeds while maintaining success rates comparable to the baseline \cite{yu2025mastering} in most settings. Notably, our method achieves peak velocities up to 7.1 m/s on lower difficulty terrains and sustains approximately 6 m/s in more complex scenarios, exhibiting superior agility compared to the $\sim$5 m/s maximum speed of \cite{yu2025mastering}. Furthermore, DA significantly enhances robustness at higher difficulty levels, particularly by bridging the dynamics gap.




\begin{table}[h!]
\renewcommand\arraystretch{1.1}
\setlength{\tabcolsep}{1.2pt} 
\caption{Sim-to-sim Performance}
\label{sim_performance}
\centering
\scriptsize 
\newcolumntype{Y}{>{\centering\arraybackslash}X}

\begin{tabularx}{\linewidth}{c|c|Y Y|Y Y|Y Y}
\toprule
\multirow{2}{*}{\textbf{Method}}
& \multirow{2}{*}{\makecell{\textbf{Difficulty} \\ \textbf{Level}}}
& \multicolumn{2}{c|}{\textbf{Zigzag}} 
& \multicolumn{2}{c|}{\textbf{Circular}} 
& \multicolumn{2}{c}{\textbf{Ellipse}} \\
& & \textbf{SR} & $\mathbf{v_{max}}$ & \textbf{SR} & $\mathbf{v_{max}}$ & \textbf{SR} & $\mathbf{v_{max}}$ \\
\midrule
\multirow{4}{*}{\makecell{Ours \\ w/ \text{DA}}} 
& 0 & 10/10 & 7.1 & 10/10 & 6.4 & 10/10 & 7.1 \\
& 3 & 10/10 & 7.0 & 10/10 & 6.2 & 10/10 & 6.4 \\
& 6 & 10/10 & 6.1 & 10/10 & 5.8 & 9/10 & 6.3 \\
& 9 & 7/10 & 6.1 & 9/10 & 5.6 & 7/10 & 5.9\\
\midrule
\multirow{4}{*}{\makecell{Ours \\ w/o \\ \text{DA}}} 
& 0 & 10/10 & 6.8 & 10/10 & 5.9 & 10/10 & 6.5 \\
& 3 & 10/10 & 6.9 & 10/10 & 6.0 & 10/10 & 5.7 \\
& 6 & 8/10 & 6.1 & 7/10 & 5.7 & 7/10 & 6.2 \\
& 9 & 2/10 & 5.9 & 5/10 & 5.2 & 3/10 & 5.6\\
\midrule
\multirow{4}{*}{\cite{yu2025mastering}} 
& 0 & 10/10 & 5.4 & 10/10 & 5.0 & 10/10 & 5.2 \\
& 3 & 10/10 & 5.2 & 10/10 & 4.9 & 10/10 & 5.4 \\
& 6 & 10/10 & 5.2 & 10/10 & 4.9 & 10/10 & 5.3 \\
& 9 & 6/10 & 4.7 & 10/10 & 4.7 & 10/10 & 5.2\\
\bottomrule
\end{tabularx}

\vspace{0.5em}
\scriptsize  $v_{max}$ denotes Maximum Velocity (m/s), \text{SR} denotes success rate, and \text{DA} denotes Delta Action Model.
\end{table}

\begin{figure}
    \centering
    \includegraphics[width=1\linewidth]{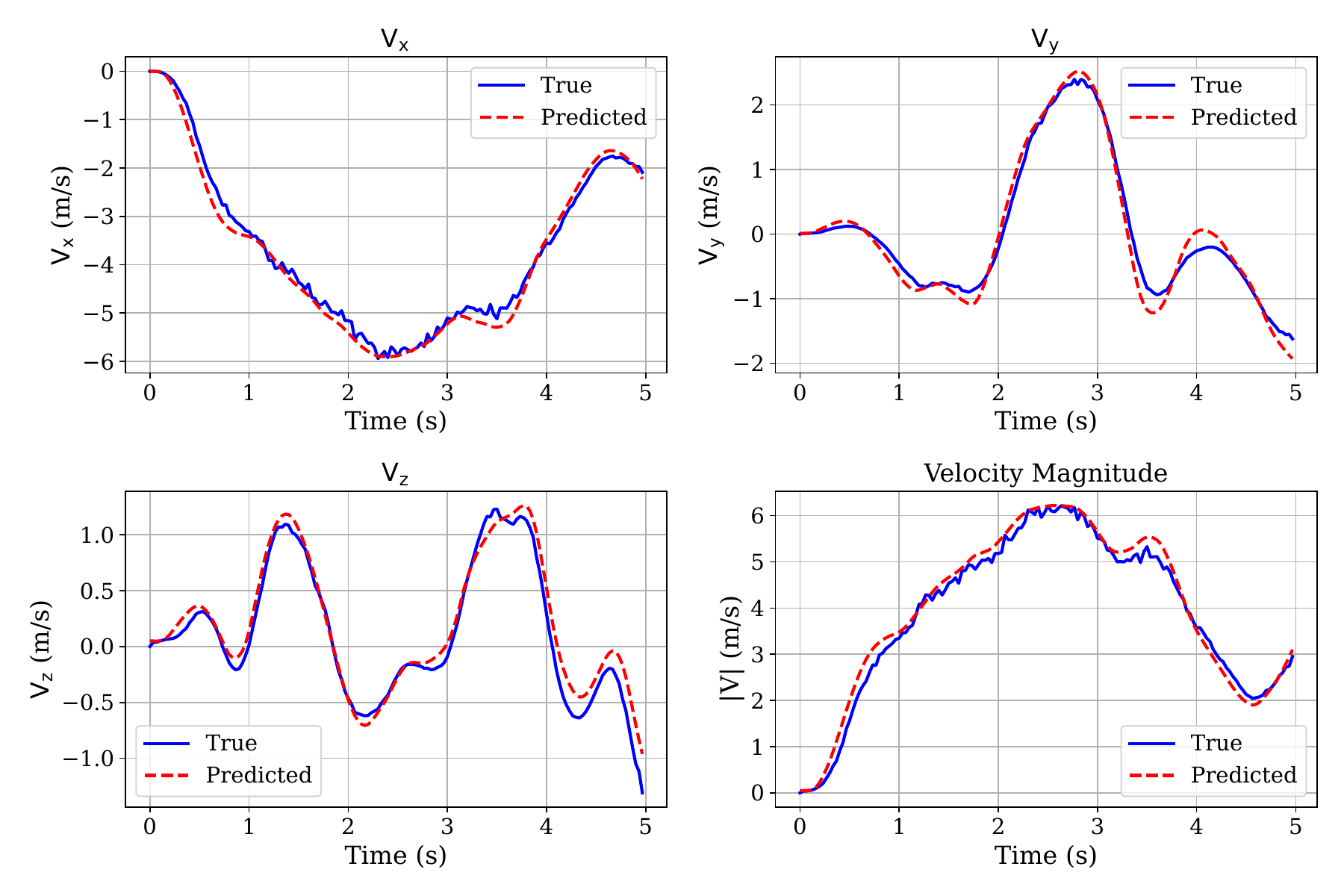}
    \caption{Velocity components during real-world flight. The dashed line shows the velocity predicted by the simulation with Delta Action Model compensation, while the solid line represents the velocity measured by the motion capture system.}
    \label{speed_component}
\end{figure}

\subsection{Real-world Experiments}

\begin{figure}
    \centering
    \includegraphics[width=0.95\linewidth]{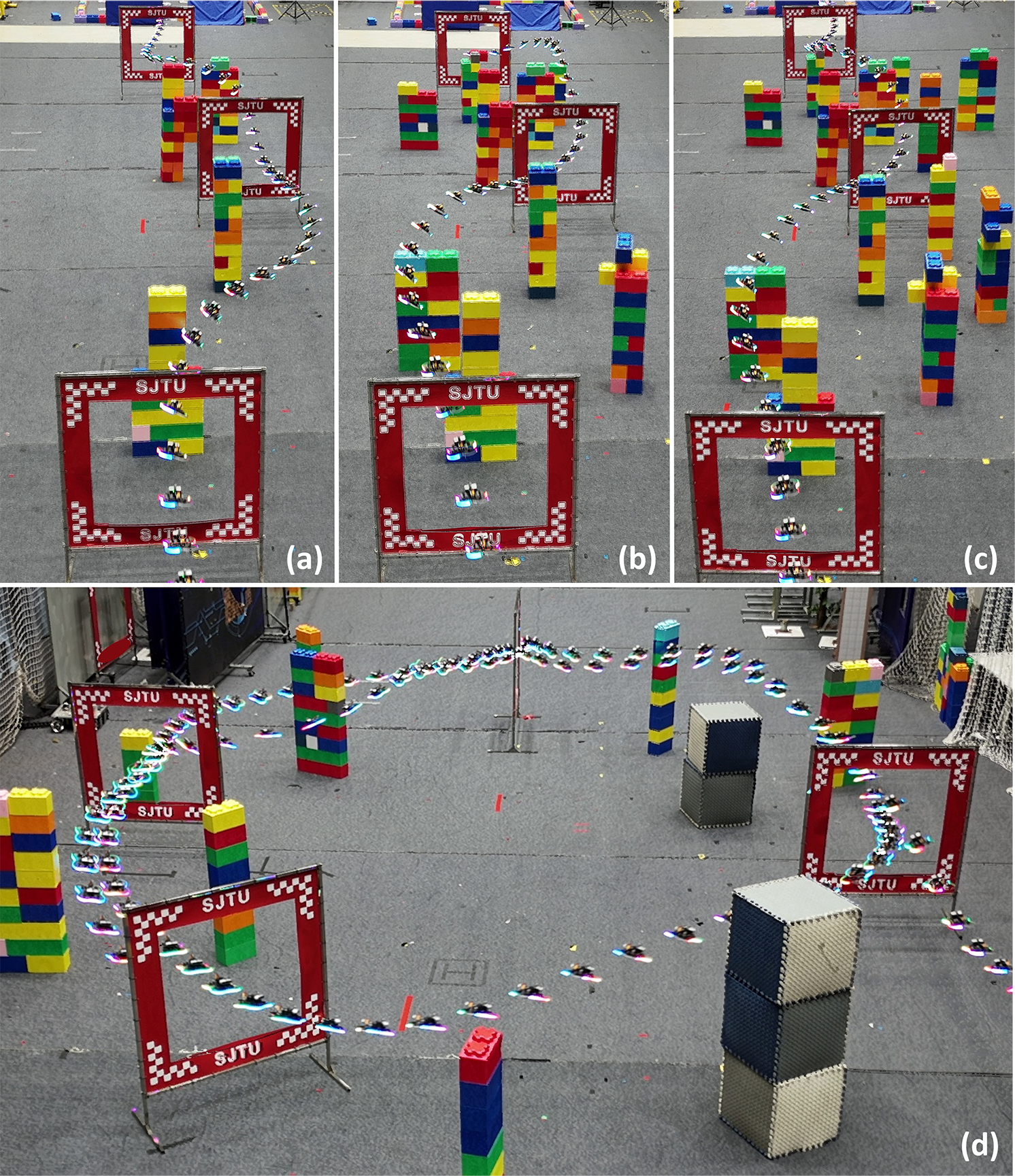}
    \caption{Real-world experiments. The policy is deployed on a physical drone navigating (a-c) zigzag and (d) circular tracks with complex, obstacle-dense layouts that were unseen during training. The maximum velocities recorded in (a), (b), and (c) are 5.9, 6.4, and 6.0 m/s, respectively.}
    \label{fig:exp_real}
\end{figure}

To validate the effectiveness of our proposed DiffRacing framework, we conducted real-world experiments on zigzag (Fig. \ref{fig:exp_real}(a-c)) and circular tracks (Fig. \ref{fig:exp_real}(d)). Crucially, the explicit track layouts and obstacle distributions in these trials were unseen during training; the policy network received only the relative position of the next gate. The system is implemented on the Betaflight\footnote{https://www.betaflight.com/} flight controller and runs on the Radxa Zero3W\footnote{https://radxa.com/products/zeros/zero3w} platform. State information, including position, orientation, and body velocity, was obtained from a motion capture system\footnote{https://www.vicon.com/}, while depth data was captured using an Intel D435i camera. Thanks to the computationally efficient policy, the network operates at 30 Hz, providing the low-latency responses necessary for adaptive navigation around unknown obstacles. To enhance sim-to-real transfer reliability, particularly in high-speed flight, we implemented the Delta Action Model in our real-world experiments. One of the resulting aligned trajectories is shown in Fig. \ref{speed_component}. As obstacle density increased, the drone consistently exhibited agile flight, reaching a top speed of 6.4 m/s (Fig. \ref{fig:exp_real}(a-c)). Additionally, continuous two-lap flights through obstacles (Fig. \ref{fig:exp_real}(d)) demonstrate the robustness of our policy in maintaining safe and sustained flight under challenging real-world conditions.


\section{Conclusion And Limitations}

In this paper, we presented DiffRacing, a novel vector field-augmented differentiable policy learning framework for vision-based autonomous drone racing, effectively balancing high-speed gate traversal with reliable obstacle avoidance. A differentiable Delta Action Model further compensates for model mismatch and enables efficient sim-to-real transfer. Extensive simulation and real-world experiments demonstrate that DiffRacing achieves superior agility and collision-free performance in complex racing environments.

\textbf{Limitations}: The current vector field is manually designed, which may limit generality. Moreover, the augmented gradient does not correspond to an explicit objective, making theoretical stability analysis nontrivial.


%





\ifCLASSOPTIONcaptionsoff
  \newpage
\fi



%


\bibliographystyle{ieeetr}
\bibliography{reference}

%








\end{document}